# Deep Bayesian Bandits: Exploring in Online Personalized Recommendations

DALIN GUO, University of California, San Diego, USA
SOFIA IRA KTENA, Twitter, UK
FERENC HUSZAR, Twitter, UK
PRANAY KUMAR MYANA, Twitter, UK
WENZHE SHI, Twitter, UK
ALYKHAN TEJANI, Twitter, UK

Recommender systems trained in a continuous learning fashion are plagued by the feedback loop problem, also known as algorithmic bias. This causes a newly trained model to act greedily and favor items that have already been engaged by users. This behavior is particularly harmful in personalised ads recommendations, as it can also cause new campaigns to remain unexplored. Exploration aims to address this limitation by providing new information about the environment, which encompasses user preference, and can lead to higher long-term reward. In this work, we formulate a display advertising recommender as a contextual bandit and implement exploration techniques that require sampling from the posterior distribution of click-through-rates in a computationally tractable manner. Traditional large-scale deep learning models do not provide uncertainty estimates by default. We approximate these uncertainty measurements of the predictions by employing a bootstrapped model with multiple heads and dropout units. We benchmark a number of different models in an offline simulation environment using a publicly available dataset of user-ads engagements. We test our proposed deep Bayesian bandits algorithm in the offline simulation and online AB setting with large-scale production traffic, where we demonstrate a positive gain of our exploration model.

## 1 INTRODUCTION

Deep learning is widely deployed to support a range of personalization use-cases [33]: from content recommendations [8, 10], to display and performance advertising [14, 31]. To keep up with shifting user preferences [14] and deal with cold-start problems, models are often updated in a continuous training loop: repeatedly fine-tuned using recent data while a previous version was deployed to serve content. This introduces a feedback loop as training data was selected by the model, a problem known as algorithmic (or selection) bias [5]. This bias may result in unjustified amplification of certain items based on spurious patterns in the data, or a failure to explore promising candidates to users altogether.

There are two main toolkits to address algorithmic bias: causal inference [4] or bandit theory and reinforcement learning (RL) [7, 17]. A central concept is exploration: all items are presented to all users uniformly to form an unbiased dataset; however, overly exploring harms the quality of the recommendations and user satisfaction. The exploration/exploitation trade-off is naturally formulated as a (contextual) multi-armed bandit task, for which an $\epsilon$-greedy policy is a simple yet powerful approach. However, as the number of candidate items is usually large, and only a handful will yield engagement for a given user, uniform exploration can have a revenue loss whilst providing

Authors' addresses: Dalin Guo, dag082@ucsd.edu, University of California, San Diego, 9500 Gilman Dr., La Jolla, CA, USA, 92093; Sofia Ira Ktena, Twitter, 1 Thørväld Circle, London, UK, siraktena@twitter.com; Ferenc Huszar, Twitter, 1 Thørväld Circle, London, UK, fhuszar@twitter.com; Pranay Kumar Myana, Twitter, 1 Thørväld Circle, London, UK, pmyana@twitter.com; Wenzhe Shi, Twitter, 1 Thørväld Circle, London, UK, wshi@twitter.com; Alykhan Tejani, Twitter, 1 Thørväld Circle, London, UK, atejani@twitter.com.

little to no information and future expected gain. Alternative methods that are driven by uncertainty, such as Upper Confidence Bound (UCB) [3, 15] and Thompson sampling (TS) [1, 30], have been proposed with theoretical guarantees.

Applying bandit algorithms to real-world settings can be challenging, as the performance relies on accurate assumptions about the reward environment, which can be highly complex. For example, the reward rate can be a nonlinear function of the context. Besides, UCB and TS require sampling from the posterior distribution, which is computationally challenging if the value function is approximated by a deep neural network to capture non-linearity.

Deep neural networks have been successfully used to predict the click through rate (CTR) for the item candidates [8, 14] and to approximate nonlinear value functions in RL setups [21]. However, traditional neural network architectures provide point estimates without uncertainty. Bayesian deep learning [22] provides a natural solution, but it is computationally expensive and challenging to train and deploy as an online service. Other methods [12, 16, 28] have been proposed to approximate the posterior distributions or estimate model uncertainty of a neural network.

This paper is inspired by bandit algorithms and posterior approximation algorithms for deep neural networks. We focus on applying them tractably in large-scale deep learning-based recommender systems. We also propose a hybrid model that contains dropout units only in the second-to-last layer, which can also be viewed as a bootstrapped model with multiple heads and shared bottom layers. We use UCB exploration, where we numerically estimate confidence intervals. We focus our experiments on display advertising, but our method can be easily adapted to other applications.

We compare our model with baselines in an offline simulated environment to show its effectiveness, and further perform an online A/B test to demonstrate its positive impact. Our main contributions are three-fold:

(1) We propose an offline simulation environment using a public dataset, and benchmark exploration techniques.
(2) We propose an efficient deep Bayesian bandits algorithm showing significant gains offline.
(3) We present results of the proposed model in an online A/B testing experiment with large scale X data (anonymity).

## 2 RELATED WORK

### 2.1 Contextual Bandits

Traditional approaches to recommender systems commonly face the cold-start problem, where contextual bandits have emerged as a viable alternative [17, 20, 24, 35]. [17] proposes LinUCB algorithm to efficiently compute the confidence interval in closed form, which shows a better performance compared to context-free model and $\epsilon$-greedy policy. LinUCB assumes a linear payoff model [9], and is evaluated offline with logged events that was collected randomly [18]. [35] proposes a partial personalization approach that uses users' latent class structure to train a set of model parameters for each class, reducing the need on the user features to fully capture the variability. Their approach provides good recommendations for new users more quickly and yields lower regret bound.

Bart was proposed [20] to jointly optimize recommendations and associated explanations and provide more transparent suggestions to users for music recommendations. Bart is incorporated with $\epsilon$-greedy exploration for easier implementation in production and propensity scoring when jointly optimizing for items and recommendation explanations. In the personalized music recommendations space, [32] leverages UCB exploration with context.

### 2.2 Deep Reinforcement Learning

Deep neural networks provide a powerful nonlinear payoff model, while introducing challenges in terms of obtaining samples from the posterior distribution. Bootstrapped DQN adapts TS that allows temporally extended exploration through randomized value functions by approximating a distribution over $Q$-values via the bootstrap [23]. A recent work [25] performs an extensive investigation of deep Bayesian bandit methods and compare different posterior approximations from an empirical standpoint under the prism of TS, in downstream bandit tasks with simulators generated with synthetic and real-world datasets. Another recent work [27] directly compares the accuracy of predictive uncertainty under input distribution shifts.

Deep reinforcement learning approaches have also been adopted in recommender system. [19] aims to model long-term rather than immediate rewards and captures the dynamic adaptation of user preferences and the interactive nature between users and recommender systems, with an "actor-critic" structure. DeepPage [34], a page-wise recommendation framework, jointly optimizes a page of items and incorporates real-time user feedback. The latter approach is evaluated in a simulated online environment of an e-commerce product illustrating potential for a production system, while that is not the case for the former.

## 3 OUR APPROACH

Here, we formulate the ads recommendation as a contextual bandit problem, where the context contains both user and ad features. To trade-off between exploration and exploitation, we consider two exploration algorithms - UCB and Thompson sampling. To address the lack of an uncertainty estimate with neural networks, we consider using dropout, bootstrapping, and propose a hybrid method that combines the idea of dropout and bootstrapping.

We use a neural network to predict CTR and additionally use a posterior approximation algorithm to obtain the model uncertainty. Given the samples of the CTR estimates for all candidate ads, an exploration algorithm picks K items to display for the given user. The user actions for the recommended items are logged and later used to fine-tune the neural network after some fixed duration.

### 3.1 Exploration techniques

*3.1.1 $\epsilon$-greedy.* $\epsilon$-greedy does not take into account of any uncertainty estimate. It 'greedily' recommends item with the highest CTR with probability 1-$\epsilon$ and randomly selecting other items uniformly with probability $\epsilon$.

*3.1.2 Thompson Sampling (TS).* Thompson sampling [30] is also known as posterior sampling or probability matching, as it samples from the posterior distribution of CTR of each item once, and acts 'greedily' according to those samples. Thus, it selects an item with a probability that this item is optimal given the current knowledge, i.e. the probability of this item having a higher CTR than all other items.

*3.1.3 Upper Confidence Bound (UCB).* UCB [2, 3] chooses 'greedily' according to the upper confidence bound of each item. It adds an uncertainty bonus to the mean estimation, based on the principle of optimism in the face of uncertainty. Here, we numerically estimate the confidence bound by empirical CDF value estimator given the samples [11].

### 3.2 Posterior Approximation algorithms

As UCB and TS require at least samples from the posterior, we consider two previously proposed algorithms that enable us to draw samples from the posterior distribution of a neural network. Both methods are computationally costly in some ways, therefore we propose an alternative method that requires less computation.

*3.2.1 Bootstrapping.* Bootstrapping obtains the uncertainty by training multiple identical models on different subsets of the dataset [16]. It is computationally expensive: (a) During training, we need to (i) partition and store the masks of data, and (ii) train multiple neural networks. (b) During prediction, we need run the forward pass of all bootstrapped networks once (if using TS) or multiple times (if using UCB). The forward pass can be expensive in a recommender system with large dataset and/or large network, which might not meet the latency requirement of an online service.

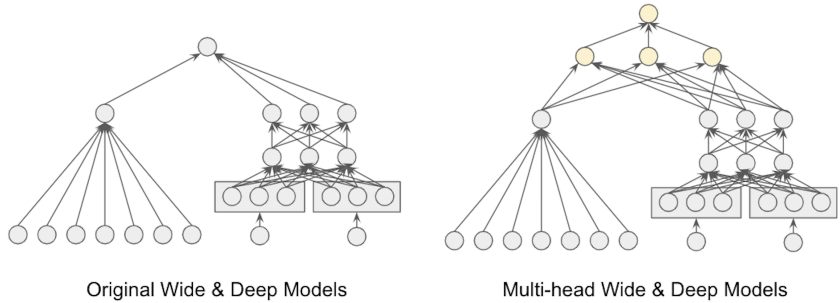

Fig. 1. Online model. Left: original wide-and-deep model. Right: Modified multihead wide-and-deep model.

To mitigate this problem, multi-head networks have been proposed [23]. This approach suggests to share the bottom (early) layers across bootstrapped networks, with each subset passing through a different "head" during training. During testing, we can obtain the estimation from all neural networks in one forward pass by taking the outputs of all the heads. However, we still need to partition the dataset. Previous studies have found that the randomness introduced by the stochastic gradient descent (SGD) optimizer and random initialization is sufficient to provide good performance in downstream tasks [23, 25]. Here, we compare all the variants of bootstrapping methods, including the original bootstrapping method (Bootstrap), using a multi-head neural network (Multihead), multiple networks trained on the same whole dataset (SGD), and multi-head network with heads trained on the whole dataset (Multihead SGD).

*3.2.2 Dropout.* Dropout during inference phase has been proposed to approximate the posterior distribution with good empirical performance and theoretical guarantees [12]. To obtain model uncertainty, the model makes predictions through multiple forward passes with different dropout units to obtain samples from the posterior distribution. Compared to bootstrapping, dropout has a lower computational cost; however, it usually takes longer to train a neural network with dropout [29], and running the forward passes multiple times can be expensive as discussed above.

*3.2.3 A Hybrid Method.* We combine the ideas of bootstrapping and dropout with the following objectives: (1) dynamically assign membership of each data point to subsets without storing the mask; (2) avoid training multiple neural networks and (3) reduce the computational cost of running multiple forward passes through the whole neural network required by the dropout method. Our model adds an additional dropout layer as the second-to-last layer. Thus, when sampling from the posterior distribution, we only need to compute the bottom layers once, and multiples pass through the dropout layer can be done in parallel without adding computational resources. The dropout layer approximately acts as the "heads" equivalent in the multi-head network, and the dropout automatically provides a Bernoulli mask for each training data point without explicitly partitioning the dataset.

### 3.3 Neural network architectures

*3.3.1 Offline simulation.* We use a fully-connected feed-forward network. The network inputs the concatenated user and ads features, while its output is a scalar value that corresponds to the score of the CTR for this user – ad pair.

*3.3.2 Online Experiment.* We use a (modified) wide-and-deep neural network [8]. In the original model (Fig. 1 left), the wide component corresponds to a generalized linear model, while the deep component corresponds to a feed-forward neural network. To adapt it for our proposed posterior approximation method, we add one additional layer that can be viewed as a multi-head layer (Fig. 1 right). The loss function used is proposed in [14].

## 4 EXPERIMENT

### 4.1 Setup

*4.1.1 Continuous Training and Self-training Setup.* Ad impressions are served to users and the label is then published to a data stream which the model's training service subscribes to. The model is warm-started from the previous version, and fine-tuned with newly collect data. In the offline simulation, the model is updated every after 20 users. In the online experiment, the continuous training process outputs models every 10 minutes to serve online traffic. The model is only trained with the data that is generated by that model, forming a self-training loop.

### 4.2 Offline Simulation.

Off-policy evaluation is challenging and the accuracy of existing approaches are reliable when the two policy are similar [13], or the logged data is random [18], which is not practical sometimes as the data was collected with pure exploitation. We first evaluate the models in a simulated environment generated by a small dataset to validate our approach, without making strong assumptions on user behavior.

*4.2.1 Dataset.* ADS-16 dataset [26] is a publicly available dataset that contains ratings of 300 ads shown to 120 users. The dataset contains full user-ad interaction matrix, which is not the case for some other popular publicly available datasets such as Criteo [6]. Each user provided a numerical ratings of how likely they will click on the ads, ranging from 1-star (negative) to 5-stars (positive). We convert the numerical ratings to binary click/no click by a threshold suggested by the original paper [26]. We extract the users and ads features, which results in 250 user features and 323 ads features. The categorical features are one-hot encoded. We randomly held-out 5 ads for each user as the test set.

*4.2.2 Metrics.* We use *area under precision-recall curve* (PR-AUC) to evaluate the trained model predictive performance. We use accumulated averaged CTR to measure the reward obtained by the model, which is negatively related to the regret that is commonly used to evaluate bandit algorithms.

*4.2.3 Hyperparameters.* The model randomly choose ads for 20 users to begin with. For each user, the model selects seven different ads. The hyperparameters used for the experiments are: RMSProp optimizer; learning rate 0.1; decay rate 0.5; batch size 64; training epochs 100; dropout rate 0.5; $\epsilon$ 0.1 (for $\epsilon$-Greedy); # samples for UCB (# of bootstrapped networks/heads): 10; 90% confidence bound: 2th largest value of ten samples. The feedforward neural network contains 2 layers with 100 and 50 units each, with an additional layer of 20 units for the hybrid method.

### 4.3 Online Experiment

*4.3.1 Dataset.* For the offline model evaluation based on X (anonymity) data, we trained on 1 day of data and test on the first hour of the following day. The training data is ∼ 550 million ads, while the testing data is ∼ 20 million.

In online experiment, models were each serving and trained on a continuous data stream of 2 % of production traffic in real time for two weeks. When computing the RCE of the trained models, we use a test data that were collected by a random policy serving 1% of production traffic (∼ 160,000), which is unbiased and representative of the distribution of the possible ad candidates at prediction time.

*4.3.2 Metrics.* We use *relative cross entropy* (RCE) to evaluate the model predictive performance. RCE measures the improvement of a model relative to the straw man, or the naive prediction, in cross entropy (CE) [14]. We also present results using *area under an receiver operating characteristic Curve* (ROC-AUC) as it is a more commonly used metric.

*4.3.3 Hyperparameters.* The hyperparameters used for the experiments on X (anonymity) data are: stochastic gradient descent (SGD) optimizer, learning rate 0.01; decay rate 0.000001; batch size 32; dropout rate: 0.5; # of samples: 100; UCB confidence bound: 5th largest value. A mix of categorical and continuous features are discretized into a fixed number of bins. The deep part of the wide-and-deep model consists of 4 layers with sizes [400, 300, 200, 100] with ReLU activation function applied for the intermediate layers. The weights are initialized using Glorot initialization.

We observed that the hybrid model tends to over-predict the CTR, as it outputs the 95th percentile score. This results in several undesired downstream consequences, interacting with other parts of the system, so we scaled down the score by multiplying with a constant to mitigate, which does not affect the ranking of the items.

## 4.4 Results

Table 1. Offline simulation: performance comparison

| Model | CTR (+%) | PR-AUC |
|---|---|---|
| Random | 0 | 0.5 |
| Greedy | 91.77 | 0.6565 |
| $\epsilon$-greedy | 91.94 | 0.6501 |
| Dropout TS | 94.60 | 0.6421 |
| Dropout UCB | 97.16 | 0.5236 |
| Bootstrap TS | 94.83 | 0.5519 |
| Bootstrap UCB | 139.03 | 0.5307 |
| SGD UCB | 127.95 | 0.5335 |
| Multihead UCB | 112.79 | 0.5279 |
| Multihead SGD UCB | 96.30 | 0.5218 |
| Hybrid TS | 67.56 | 0.6311 |
| Hybrid UCB | 82.44 | 0.5165 |

Table 2. Offline simulation: Warm-start the hybrid model

| Model (# epochs) | train PR-AUC | CTR (+%) | test PR-AUC |
|---|---|---|---|
| Random | 0.5 | 0 | 0.5 |
| $\epsilon$-greedy (100) | 0.5951 | 94.30 | 0.6692 |
| Hybrid (100) | 0.5001 | 85.99 | 0.5108 |
| Hybrid (200) | 0.5584 | 60.51 | 0.5165 |
| Hybrid (500) | 0.5895 | 128.66 | 0.5294 |

Table 3. Predictive performance of models self-trained in online A/B test

| Model | RCE | ROC-AUC(%) |
|---|---|---|
| Hybrid | **8.12** | **68.37** |
| Control | 7.95 | 67.13 |

*4.4.1 Offline evaluation.* We test the bootstrapping and dropout methods with TS and UCB in the simulated environment, and the results are shown in Table 1. We compared all models against the random policy, which chooses ads to serve uniformly from all candidate ads. We calculated the percentage of increase of CTR against the random policy, and PR-AUC of the final model on the test set. We observe a trade-off between averaged increased CTR and PR-AUC, as the model trained on more random data is expected to perform better than the model trained on the biased data. UCB in general earns more reward than TS. Bootstrap UCB earns the highest rewards, while having the highest computational cost at the same time. The performance drops as we reduce the computational cost of the Bootstrap UCB model by using other variants, including the Hybrid model we proposed.

One hypothesis that the dropout-based method, including the proposed Hybrid method, does not perform well is that a neural network with dropout units usually takes longer to converge, typically 2-3 times [29]. One potential solution is to offline warm-start the hybrid model with longer training epochs before deploy it online. To test the hypothesis, we warm-start the models with a dataset collected by a greedy policy, which is also available for an online service deployment in general. We keep the same online training epoch for online continuous training for all models. As expected, the training PR-AUC is much lower for the neural network with dropout units (Hybrid) compared to without dropout units ($\epsilon$-greedy) under same training epochs (100), shown in Table 2. As we train the dropout network longer, the training PR-AUC increases, and also the accumulated CTR and test PR-AUC. With longer offline training epochs, the hybrid model perform comparable to SGD UCB while has lower computational cost. Note that the test PR-AUC of this dropout network, given short online training epochs, is not much lower than the network without dropout (e.g. SGD UCB), suggesting that the good predictive performance of a dropout network can be maintained at least for a period of time. This results provide a potential pipeline for online deployment of the hybrid model, that the model can be warm-started offline with longer training epochs before deployment, or even periodically if the predictive accuracy starts to drop too low with limited online training epochs.

*4.4.2 Online evaluation.* To ensure the modified wide-and-deep has a similar performance as the production model, we first validate it on offline X (anonymity) data. We perform a quick hyperparameter tuning of the number of dropout units and dropout rate. The model with the best hyperparameters (25 dropout units, 0.5 dropout rate) achieves similar RCE (-0.0253) as the production model. We do not find any advantage of training the model with additional epochs.

We compare our proposed hybrid model with current production model (greedy policy) in an online A/B experiment. We find that the cost of exploration is not significant – serving +2% ads with a flat revenue (no significant +/-), and no significant decrease in training and serving speed. We observe no direct improvement in product metrics in our experiment. Note that serving more ads might not linearly result in revenue gain, as the quality of ads decreases.

We further evaluate the benefit of the information gained from exploration, i.e., a higher predictive performance. We disabled the dropout units for model evaluation. We took a snapshot of the models at time *t hour*, and evaluated on randomly served ads from *t* to *t+1* hour. The hybrid model trained with data selected by itself has a higher RCE and ROC-AUC than the production model (Table 3). It indicates that exploratory data collected by hybrid model improves the model performance, since the differences of model architectures results in a lower RCE for the hybrid model.

We also tested $\epsilon$-greedy policy in a previous online A/B experiment, which resulted in 100% increase in negative engagement rate (users negatively engaging with the ad, dismiss etc.), indicating a negative impact of $\epsilon$-greedy on user experience. With our proposed hybrid model, we observe no increase in negative engagement rate (-2%). This can be a result of the directed exploration of UCB, which only shows ads that more or less consistent with user's preferences (i.e., ads has high predictive CTR and high uncertainty), rather than randomly picking ads for exploration.

## 5 CONCLUSION

In this paper we have explored combining bandit algorithms with deep neural networks, to address the algorithmic bias issue and to optimize for long-term reward. We proposed a hybrid method that contains dropout units only in the second-to-last layer, acting as "heads" in a multihead network. We performed offline comparison and online AB testing with large scale in-house data, showing good performance of our proposed method with a lower computational cost.

We did not observe direct or immediate benefit in production metrics; however, the improvement in model performance will potentially lead to revenue gain in the long-run. In addition, we did not modify other components of the online service, such as pricing strategy. Future work can include a suitable pricing strategy for a recommender system with exploratory behavior, in which a more careful calibration of the model is needed.

## REFERENCES
[1] Shipra Agrawal and Navin Goyal. 2013. Thompson sampling for contextual bandits with linear payoffs. In *International Conference on Machine Learning*. 127–135.
[2] Peter Auer. 2002. Using confidence bounds for exploitation-exploration trade-offs. *Journal of Machine Learning Research* 3, Nov (2002), 397–422.
[3] Peter Auer, Nicolo Cesa-Bianchi, and Paul Fischer. 2002. Finite-time analysis of the multiarmed bandit problem. *Machine learning* 47, 2-3 (2002), 235–256.
[4] Léon Bottou, Jonas Peters, Joaquin Quiñonero-Candela, Denis X Charles, D Max Chickering, Elon Portugaly, Dipankar Ray, Patrice Simard, and Ed Snelson. 2013. Counterfactual reasoning and learning systems: The example of computational advertising. *The Journal of Machine Learning Research* 14, 1 (2013), 3207–3260.
[5] Allison JB Chaney, Brandon M Stewart, and Barbara E Engelhardt. 2018. How algorithmic confounding in recommendation systems increases homogeneity and decreases utility. In *Proceedings of the 12th ACM Conference on Recommender Systems*. 224–232.
[6] Olivier Chapelle. 2014. Modeling delayed feedback in display advertising. In *Proceedings of the 20th ACM SIGKDD international conference on Knowledge discovery and data mining*. ACM, 1097–1105.
[7] Minmin Chen, Alex Beutel, Paul Covington, Sagar Jain, Francois Belletti, and Ed H Chi. 2019. Top-k off-policy correction for a REINFORCE recommender system. In *Proceedings of the Twelfth ACM International Conference on Web Search and Data Mining*. 456–464.
[8] Heng-Tze Cheng, Levent Koc, Jeremiah Harmsen, Tal Shaked, Tushar Chandra, Hrishi Aradhye, Glen Anderson, Greg Corrado, Wei Chai, Mustafa Ispir, et al. 2016. Wide & deep learning for recommender systems. In *Proceedings of the 1st workshop on deep learning for recommender systems*. ACM, 7–10.
[9] Wei Chu, Lihong Li, Lev Reyzin, and Robert Schapire. 2011. Contextual bandits with linear payoff functions. In *Proceedings of the Fourteenth International Conference on Artificial Intelligence and Statistics*. 208–214.
[10] Paul Covington, Jay Adams, and Emre Sargin. 2016. Deep neural networks for youtube recommendations. In *Proceedings of the 10th ACM conference on recommender systems*. 191–198.
[11] Terry Dielman, Cynthia Lowry, and Roger Pfaffenberger. 1994. A comparison of quantile estimators. *Communications in Statistics-Simulation and Computation* 23, 2 (1994), 355–371.
[12] Yarin Gal and Zoubin Ghahramani. 2016. Dropout as a bayesian approximation: Representing model uncertainty in deep learning. In *international conference on machine learning*. 1050–1059.
[13] Alexandre Gilotte, Clément Calauzènes, Thomas Nedelec, Alexandre Abraham, and Simon Dollé. 2018. Offline a/b testing for recommender systems. In *Proceedings of the Eleventh ACM International Conference on Web Search and Data Mining*. 198–206.
[14] Sofia Ira Ktena, Alykhan Tejani, Lucas Theis, Pranay Kumar Myana, Deepak Dilipkumar, Ferenc Huszar, Steven Yoo, and Wenzhe Shi. 2019. Addressing delayed feedback for continuous training with neural networks in CTR prediction. In *Proceedings of the 13th ACM Conference on Recommender Systems*. ACM, 187–195.
[15] Tze Leung Lai and Herbert Robbins. 1985. Asymptotically efficient adaptive allocation rules. *Advances in applied mathematics* 6, 1 (1985), 4–22.
[16] Balaji Lakshminarayanan, Alexander Pritzel, and Charles Blundell. 2017. Simple and scalable predictive uncertainty estimation using deep ensembles. In *Advances in neural information processing systems*. 6402–6413.
[17] Lihong Li, Wei Chu, John Langford, and Robert E Schapire. 2010. A contextual-bandit approach to personalized news article recommendation. In *Proceedings of the 19th international conference on World wide web*. 661–670.
[18] Lihong Li, Wei Chu, John Langford, and Xuanhui Wang. 2011. Unbiased offline evaluation of contextual-bandit-based news article recommendation algorithms. In *Proceedings of the fourth ACM international conference on Web search and data mining*. 297–306.
[19] Feng Liu, Ruiming Tang, Xutao Li, Weinan Zhang, Yunming Ye, Haokun Chen, Huifeng Guo, and Yuzhou Zhang. 2018. Deep reinforcement learning based recommendation with explicit user-item interactions modeling. *arXiv preprint arXiv:1810.12027* (2018).
[20] James McInerney, Benjamin Lacker, Samantha Hansen, Karl Higley, Hugues Bouchard, Alois Gruson, and Rishabh Mehrotra. 2018. Explore, exploit, and explain: personalizing explainable recommendations with bandits. In *Proceedings of the 12th ACM Conference on Recommender Systems*. 31–39.
[21] Volodymyr Mnih, Koray Kavukcuoglu, David Silver, Andrei A Rusu, Joel Veness, Marc G Bellemare, Alex Graves, Martin Riedmiller, Andreas K Fidjeland, Georg Ostrovski, et al. 2015. Human-level control through deep reinforcement learning. *Nature* 518, 7540 (2015), 529.
[22] RM Neal. 1995. Bayesian learning for neural networks [PhD thesis]. *Toronto, Ontario, Canada: Department of Computer Science, University of Toronto* (1995).
[23] Ian Osband, Charles Blundell, Alexander Pritzel, and Benjamin Van Roy. 2016. Deep exploration via bootstrapped DQN. In *Advances in neural information processing systems*. 4026–4034.
[24] Filip Radlinski, Robert Kleinberg, and Thorsten Joachims. 2008. Learning diverse rankings with multi-armed bandits. In *Proceedings of the 25th international conference on Machine learning*. 784–791.
[25] Carlos Riquelme, George Tucker, and Jasper Snoek. 2018. Deep Bayesian Bandits Showdown: An Empirical Comparison of Bayesian Deep Networks for Thompson Sampling. *International Conference on Learning Representations, ICLR*.
[26] Giorgio Roffo and Alessandro Vinciarelli. 2016. Personality in computational advertising: A benchmark. In *4 th Workshop on Emotions and Personality in Personalized Systems (EMPIRE)*.

[27] Jasper Snoek, Yaniv Ovadia, Emily Fertig, Balaji Lakshminarayanan, Sebastian Nowozin, D Sculley, Joshua Dillon, Jie Ren, and Zachary Nado. 2019. Can you trust your model's uncertainty? Evaluating predictive uncertainty under dataset shift. In *Advances in Neural Information Processing Systems*. 13969–13980.

[28] Jasper Snoek, Oren Rippel, Kevin Swersky, Ryan Kiros, Nadathur Satish, Narayanan Sundaram, Mostofa Patwary, Mr Prabhat, and Ryan Adams. 2015. Scalable bayesian optimization using deep neural networks. In *International conference on machine learning*. 2171–2180.

[29] Nitish Srivastava, Geoffrey Hinton, Alex Krizhevsky, Ilya Sutskever, and Ruslan Salakhutdinov. 2014. Dropout: a simple way to prevent neural networks from overfitting. *The journal of machine learning research* 15, 1 (2014), 1929–1958.

[30] William R Thompson. 1933. On the likelihood that one unknown probability exceeds another in view of the evidence of two samples. *Biometrika* 25, 3/4 (1933), 285–294.

[31] Ruoxi Wang, Bin Fu, Gang Fu, and Mingliang Wang. 2017. Deep & cross network for ad click predictions. In *Proceedings of the ADKDD'17*. 1–7.

[32] Xinxi Wang, Yi Wang, David Hsu, and Ye Wang. 2014. Exploration in interactive personalized music recommendation: a reinforcement learning approach. *ACM Transactions on Multimedia Computing, Communications, and Applications (TOMM)* 11, 1 (2014), 1–22.

[33] Shuai Zhang, Lina Yao, Aixin Sun, and Yi Tay. 2019. Deep learning based recommender system: A survey and new perspectives. *ACM Computing Surveys (CSUR)* 52, 1 (2019), 1–38.

[34] Xiangyu Zhao, Long Xia, Liang Zhang, Zhuoye Ding, Dawei Yin, and Jiliang Tang. 2018. Deep reinforcement learning for page-wise recommendations. In *Proceedings of the 12th ACM Conference on Recommender Systems*. 95–103.

[35] Li Zhou and Emma Brunskill. 2016. Latent contextual bandits and their application to personalized recommendations for new users. In *Proceedings of the Twenty-Fifth International Joint Conference on Artificial Intelligence*. 3646–3653.